\documentclass[letterpaper]{article} 
\usepackage{aaai25}  
\usepackage{times}  
\usepackage{helvet}  
\usepackage{courier}  
\usepackage[hyphens]{url}  
\usepackage{graphicx} 
\usepackage{natbib}  
\usepackage{caption} 
\usepackage{algorithm}

\usepackage{amsmath}
\usepackage{newfloat}
\usepackage{listings}
\usepackage{multirow}
\usepackage{multicol}
\usepackage{makecell} 
\usepackage{mathrsfs}
\usepackage{url}            
\usepackage{booktabs}       
\usepackage{amsfonts}       
\usepackage{nicefrac}       
\usepackage{microtype}      
\usepackage{xcolor}         
\usepackage{graphicx}  
\usepackage{colortbl}  
\usepackage{array}   
\usepackage{wasysym}
\usepackage{pifont}
\usepackage{amssymb}  
\usepackage{algpseudocode}
\DeclareCaptionStyle{ruled}{labelfont=normalfont,labelsep=colon,strut=off} 
\floatstyle{ruled}
\newfloat{listing}{tb}{lst}{}
\floatname{listing}{Listing}
\title{EventZoom: A Progressive Approach to Event-Based Data Augmentation for Enhanced Neuromorphic Vision}
\author{
	Yiting Dong\textsuperscript{\rm 1,3,4}\equalcontrib,  Xiang He\textsuperscript{\rm 2,3}\equalcontrib, Guobin Shen\textsuperscript{\rm 1,3,4} ,
	\textbf{Dongcheng Zhao\textsuperscript{\rm 3,4}, Yang Li\textsuperscript{\rm 2,3}, Yi Zeng\textsuperscript{\rm 1,2,3,4,5}\footnote{Corresponding Author}}\\  
}
\affiliations{
	\textsuperscript{\rm 1}School of Future Technology, University of Chinese Academy of Sciences \\
	\textsuperscript{\rm 2}School of Artificial Intelligence, University of Chinese Academy of Sciences \\
	\textsuperscript{\rm 3}Brain-inspired Cognitive Intelligence Lab, Institute of Automation, Chinese Academy of Sciences\\  
	\textsuperscript{\rm 4}Center for Long-term Artificial Intelligence\\
	\textsuperscript{\rm 5}Key Laboratory of Brain Cognition and Brain-inspired Intelligence Technology, CAS \\
	\{dongyiting2020, hexiang2021,shenguobin2021,\\
	zhaodongcheng2016,liyang2019,yi.zeng\}@ia.ac.cn \\
	
}

\usepackage{bibentry}

\begin{document}

\maketitle

\begin{abstract}
Dynamic Vision Sensors (DVS) capture event data with high temporal resolution and low power consumption, presenting a more efficient solution for visual processing in dynamic and real-time scenarios compared to conventional video capture methods. Event data augmentation serves as an essential method for overcoming the limitation of scale and diversity in event datasets.   Our comparative experiments demonstrate that the two factors, spatial integrity and temporal continuity, can significantly affect the capacity of event data augmentation, which guarantee the maintenance of the sparsity and high dynamic range characteristics unique to event data. However, existing augmentation methods often neglect the preservation of spatial integrity and temporal continuity. To address this, we developed a novel event data augmentation strategy \texttt{EventZoom}, which employs a temporal progressive strategy, embedding transformed samples into the original samples through progressive scaling and shifting. The scaling process avoids the spatial information loss associated with cropping, while the progressive strategy prevents interruptions or abrupt changes in temporal information. We validated \texttt{EventZoom} across various supervised learning frameworks.  The experimental results show that \texttt{EventZoom} consistently outperforms existing event data augmentation methods with SOTA performance. For the first time, we have concurrently employed Semi-supervised and Unsupervised learning to verify  feasibility on event augmentation algorithms, demonstrating the applicability and effectiveness of \texttt{EventZoom} as a powerful event-based data augmentation tool in handling real-world scenes with high dynamics and variability environments.

\end{abstract}

%
\begin{figure}
	\centering
	\centerline{\includegraphics[width=0.95\columnwidth]{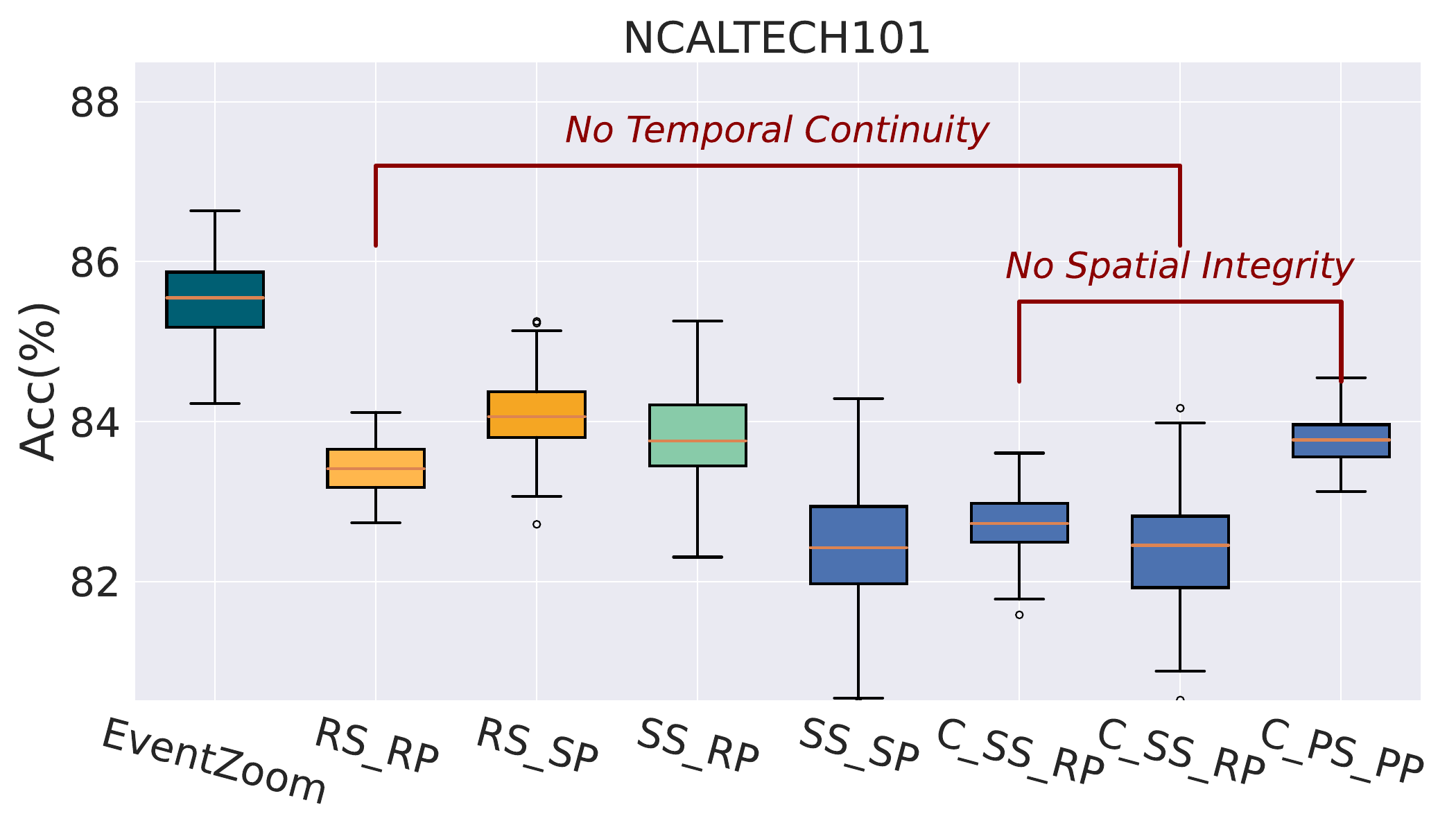} }
	\caption{Visualization of Boxplots of multiple experiments using event augmentation methods in settings lacking temporal continuity and spatial integrity.}
	\label{fig:different_setting}
\end{figure}
\section{Introduction}

Event data, as captured by Dynamic Vision Sensors (DVS), signifies a paradigm shift from traditional frame-based video capture to more sophisticated event-based or neuromorphic vision systems. \cite{hu_v2e_nodate,gallego_event_based_2022,lakshmi2019neuromorphic,taverni2018front}.  This mechanism generates asynchronous data streams and embodies several critical advantages, including high temporal resolution, low energy requirements, and substantial reductions in data redundancy\cite{schuman2017survey}. This non-continuous, and asynchronous data capture method provides a finer granularity in capturing dynamic real-world changes. It offers considerable potential for applications such as visual navigation \cite{barranco2016dataset,zujevs2021event}, autonomous driving \cite{chen2020event}, and gesture recognition\cite{amir2017low,zhang2021event}, by facilitating enhanced real-time decisions and perception capabilities.


\begin{figure*}
	\centering
	\includegraphics[width=1.95\columnwidth]{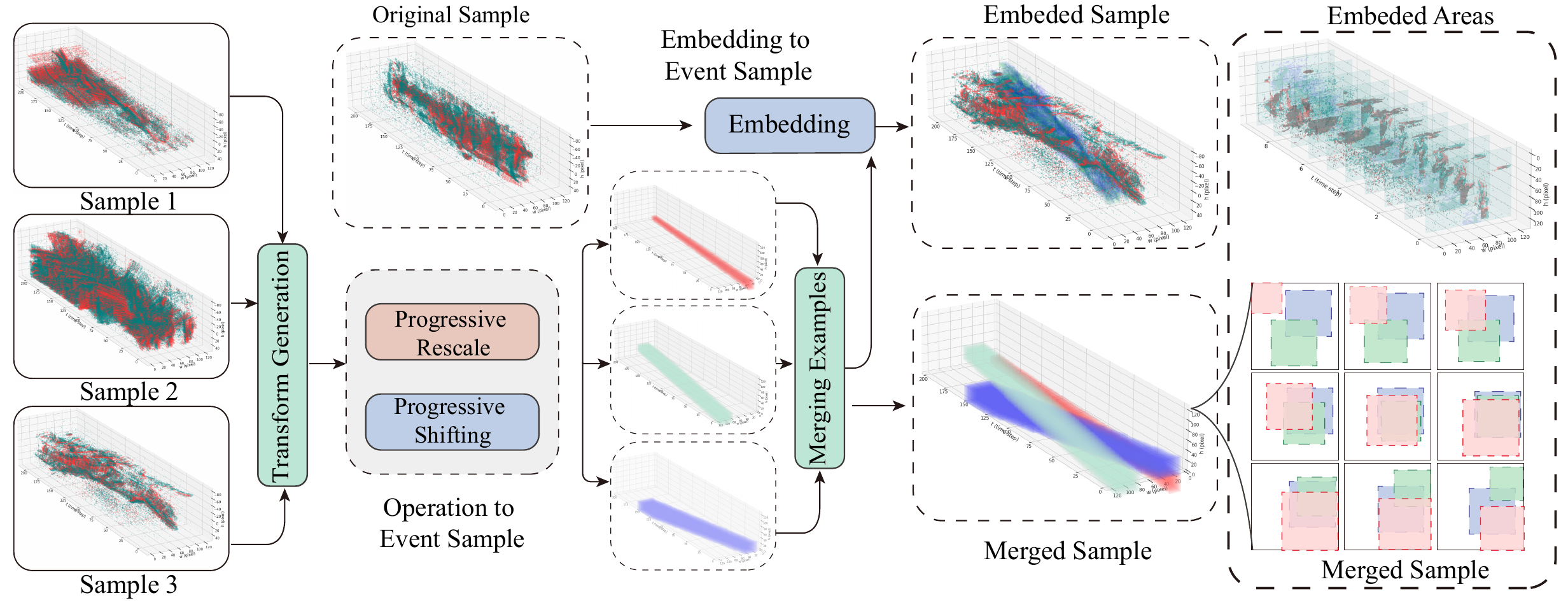}
	\caption{The data augmentation process for \texttt{EventZoom} is illustrated. \textit{Event Sample 1-3} undergoes a progressive scaling and shifting along the temporal dimension. The scaled sample are then incorporated into \textit{Original Sample }. Depending on the \textit{mixnum} settings, the number of samples varies. Each time step is assigned a unique label, which is synthesized based on the proportion of events inserted.}
	\label{fig:aug}
\end{figure*}

However, the deployment of event-based data systems in practical applications encounters significant challenges, predominantly due to the constraints on dataset availability and diversity and the inherent characteristics of event data itself\cite{gallego_event_based_2022}. Most event datasets are limited in scale and scope. The dependency on specialized event camera hardware and specific environmental conditions for data collection complicates the creation of extensive and diverse datasets.  Additionally, the inherent sparsity and the non-uniform temporal distribution of event data introduce complexities in data processing \cite{he2020comparing}.  In this context, exploring data augmentation techniques specifically designed for event data becomes crucial. These techniques offer significant potential for better utilization of this novel type of visual data.

Exploration of data augmentation techniques for event data focuses on adapting traditional augmentation methods \cite{zhang_mixup_2018,yun_cutmix_2019,devries_improved_2017} to the temporal dimension. The few existing augmentation strategies typically draw from CutMix\cite{yun_cutmix_2019} techniques, selecting and replacing parts of an event stream with another. However, we believe that the temporal continuity and spatial integrity are more critical for event data, and methods involving cutting and replacement disrupt these characteristics.  We conducted ablation experiments on the neuromporphic datasets to compare the effects under different settings to validate our hypothesis. The results, depicted in Figure \ref{fig:different_setting}, illustrate that event augmentation techniques that the lack of temporal continuity and spatial integrity detrimentally affect model's performance. This issue arises traditional augmentation methods are tailored for image data characterized by regular spatial structures \cite{li2022neuromorphic,shen2023eventmix}. Event data, which fundamentally captures dynamic lighting changes rather than static images, challenges the direct transfer of these methods, potentially disrupting the inherent temporal and spatial linkages within the data\cite{shorten_survey_2019}. 

Inspired by these insights, we developed  \texttt{EventZoom}— a data augmentation strategy that preserves both temporal continuity and spatial integrity. \texttt{EventZoom} synthesizes new event sequences using a carefully designed yet efficient algorithm. Figure \ref{fig:aug} illustrates the flow of data processing for \texttt{EventZoom}. It scales randomly event sequences and embeds them into another sample, ensuring the preservation of spatial information. Simultaneously, the embedded event sequences undergo \textbf{Progressive} \textbf{Scaling} and \textbf{Shifting} along the temporal dimension to emphasize the temporal continuity. This approach not only maintains the authenticity of the data but also enriches its diversity and complexity. \texttt{EventZoom} is distinguished by its focus on retaining detailed information in the temporal and spatial dimensions, and thus broadening the scope of the training data.

Beyond supervised learning, the effectiveness of augmentation techniques in semi-supervised and unsupervised settings serves as a valuable benchmark for assessing the strength of data augmentation strategies, as these methods rely on exploring data consistency and latent features to facilitate learning. After extensive experiments, we have validated the applicability of \texttt{EventZoom}  across various learning environments, showcasing superior ability to enhances model predictability and processing capabilities in semi-supervised and unsupervised learning. \texttt{EventZoom}  significantly enable more effective utilization of the unique advantages of event data to support the processing of real-world scenarios characterized by high dynamics and variability.

\section{Related Work}

\paragraph{Data Augmentation} Data Augmentation applied to  expands the size and diversity of training datasets. Except traditional data augmentation for image data (geometric and photometric transformations \cite{maharana_review_2022,shorten2019survey,mumuni2022data}), several advanced data augmentation techniques have emerged. Mixup \cite{mumuni2022data} technique blends features and labels from two or more images to create new samples. Cutout \cite{devries_improved_2017} randomly masks parts of the input image. CutMix \cite{yun_cutmix_2019} combines elements of Mixup and Cutout by replacing part of an image with a segment from another and appropriately blending their labels. Techniques like PuzzleMix\cite{kim2020puzzle} and SaliencyMix\cite{uddin2020saliencymix} further strategically utilize salient regions from different images to enhance the training process. Moreover, automated methods such as AutoAugment\cite{cubuk2019autoaugment} and RandAugment\cite{cubuk2020randaugment} employ reinforcement learning or random searches to discover the most effective augmentation strategies.

\paragraph{Event Data Augmentation}
The augmentation techniques for event data are specifically designed for their unique attributes, closely tied to time. EventDrop\cite{gu2021eventdrop} enhances the regularization ability of the model by randomly dropping events. NDA \cite{li2022neuromorphic} by analyzing the impact of various traditional data augmentation methods on event data and applying them accordingly. EventMix\cite{shen2023eventmix} randomly samples a Gaussian distribution in the event stream and replaces the area with a corresponding area from a random sample. EventRPG\cite{sun2024eventrpg} utilizes CAM and region salience detection techniques to correlate the cut-out area with significant regions. ShapeAug\cite{bendig2024shapeaug} achieves data augmentation by moving simple shapes within the image. EventAugment \cite{gu2024eventaugment} searches for optimal combinations of enhancements through automatic parameter tuning. However, our experiments demonstrate the importance of preserving complete spatial and temporal information for enhancing event data, which are not addressed  by the current works.

\paragraph{Data Augmentation in Semi(Un)-supervised}
Data augmentation is essential in semi-supervised \cite{berthelot2019mixmatch,sohn2020fixmatch,berthelot2019remixmatch,xie2020unsupervised} and unsupervised learning scenarios\cite{wu2018unsupervised,he2020momentum,chen2020simple,chen2021exploring}. Techniques such as consistency regularization \cite{xie2020unsupervised} rely heavily on data augmentation. They demand that models maintain consistent predictions across different augmented versions of the same data, thereby boosting the reliability of unlabeled data distribution. In the context of unsupervised learning \cite{chen2020simple,chen2021exploring}, data augmentation is instrumental in learning multiple perspectives of samples. By subjecting data to various transformations, models can discern the intrinsic structures and variations within the data, which is vital for tasks like contrastive learning. 

We provide a more comprehensive and detailed related work in Appendix A, where we thoroughly cover the background and existing research on data augmentation.

\section{Methods}

\begin{figure}[t]
	\centering
	\includegraphics[width=0.95\columnwidth]{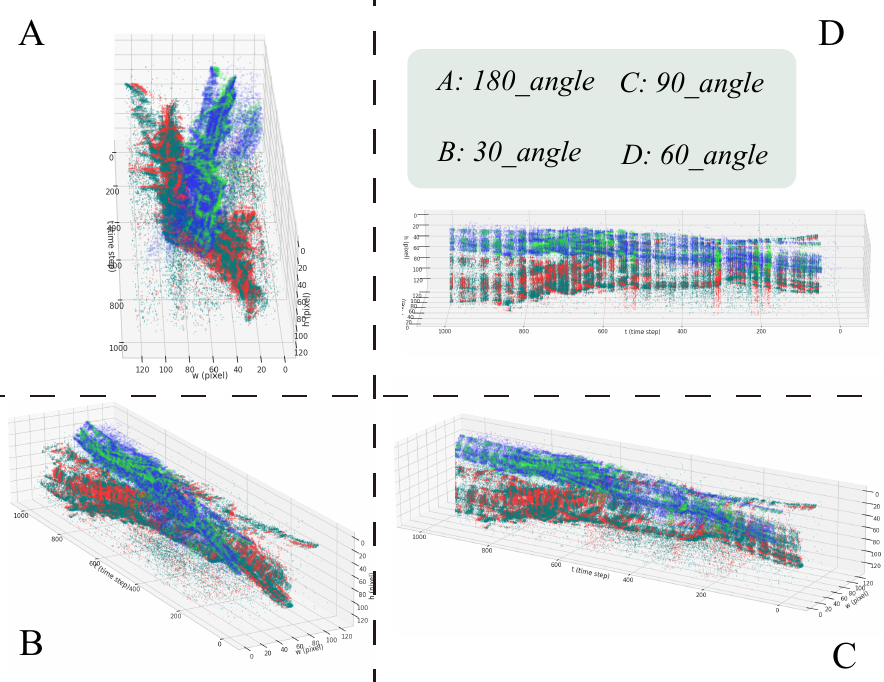} 
	\caption{Visualization of the samples generated by \texttt{EventZoom}. The zoomed sample is embeded into the original sample.}
	\label{fig:example_angle}
\end{figure}

To address the challenge of enhancing the \textit{temporal continuity} and \textit{spatial integrity} of event data, we developed a novel data augmentation strategy called \texttt{EventZoom}. This approach synthesizes new event sequences, sharing a concept akin to mixed-sample techniques. We show an event sequence generated using \texttt{EventZoom} in Figure \ref{fig:example_angle} with different views to illustrate our approach visually.
\subsection{EventZoom }

Let $x_{ori}$ represent an event sequence with dimensions $x_{ori}\in \mathbb{R}^{T \times C \times H \times W}$. $T$ denotes the time steps,  $C,W,H$ represent the channel, width, height, respectively, and $y_{ori}$ is the corresponding label. \texttt{EventZoom} augments data by randomly selecting $mixnum$ event sequences $X = \{x^i\}^{mixnum}_{i=1}$, which is not confined to the same category $y_{ori}$ but different categories $Y = \{y^i\}^{n}_{i=1}$ , $n$ means the total number of categories. 
The generation of new samples can be summarized in Equation \ref{eq:1}:
\begin{equation}
	x_{new}=F_{eventzoom}(x_{ori}, \{x^i\}^{mixnum}_{i=1}) 
	\label{eq:1}
\end{equation}
The new samples $x_{new}^i$ are generated by sequentially embedding each sample $\{x_i\}_{i=1}^{mixnum}$ into the original sample $x_{ori}$, as shown in Equation \ref{eq:2}.
\begin{equation}
	\begin{aligned}
		x_{new}^i =& G_{embed}(x_{new}^{i-1},x^i) \\
		x_{new}^0 =& x_{ori} ,\;when\; i=1
	\end{aligned}
	\label{eq:2}
\end{equation}
After each embedding, the resulting mixed sample is denoted as $x_{new}^i$. The function $F_{eventzoom}$ represents the data augmentation method, while $G_{embed}$ signifies the embedding operation performed at each time. This process is performed $mixnum$ times.

To preserve spatial integrity, the selected sequence is scaled by a random factor $\lambda$, sampled from a Beta distribution $  Beta (\lambda_{ min }, \lambda_{ max }) $. The parameters $ \lambda_{  min }$ and $ \lambda_{ max } $ define the bounds of the distribution. In practice, to achieve gradual changes, \texttt{EventZoom} randomly selects two values, \(\lambda_s\) and \(\lambda_e\), at the first and last time steps, respectively. The intermediate $\lambda$ values are obtained through linear interpolation: 
\begin{equation}
	\lambda_t = t \cdot \lambda_s + (1 - t) \cdot \lambda_e
	\label{eq:3}
\end{equation}
where $t$ represents the $t$-th time step.

To maintain the temporal continuity of the sample, the embedded sample undergoes a gradual shifting within the image boundaries. The position coordinates are determined by random values $rx$ and $ry$, which are within the ranges $(0, W)$ and $(0, H)$, respectively, where $W$ and $H$ are the width and height of the image. Similarly, to achieve gradual changes, \texttt{EventZoom} selects two random values $(rx_s, ry_s)$ and $(rx_e, ry_e)$ at the first and last time steps. The intermediate position $(rx_t, ry_t)$ is obtained through linear interpolation: 

\begin{equation}
	(rx_t, ry_t) = t \cdot (rx_s, ry_s) + (1 - t) \cdot (rx_e, ry_e)
	\label{eq:4}
\end{equation} 
The formal definition of sample mixing is given by Equation \ref{eq:5} as follows.

\begin{equation}
	G_{embed}(x_{new}^{i-1},x^i)
	=(1-M^i) \odot x_{new}^{i-1} + Zoom(x^i)
	\label{eq:5}
\end{equation}  
where
\begin{equation}
	Zoom(x^i_t)=Scale(Shift(x^i_t,(rx_t,ry_t)),\lambda_t)
	\label{eq:6}
\end{equation}  
The Equation \ref{eq:5} illustrates the operation of $G_{embed}$, where $x_i$ is embedded into the previously synthesized sample $x_{new}^{i-1}$ after applying the $Zoom$ operation. The corresponding positions in $x_{new}^{i-1}$ are then set to zero. Here, $M^i$ represents the positions for placing the patches, and $\odot$ denotes pixel-wise multiplication. The $Zoom$ operation involves a progressive scaling and shifting of the  $x^i$. Specifically, the  $Zoom$ operation scaling and shifting the $x^i_t$ at time $t$ based on the given parameters.

It is important to note that while each individual embedding involves unidirectional progressive shifting and scaling, the combination of multiple sample embeddings results in an embedded sample with varying directional changes.

For label generation, to ensure simplicity and maintain the relevance of enhanced data to the original labels, the labels are weighted according to the percentage of the sample covered by $ M $. Due to the progressive nature of the process, this percentage varies at different time steps. This can be formally defined as:
\begin{equation}
	y_{new}^i=(1-a^i_t)·y_{new}^{i-1}+a^i_t·y^i 
	\label{eq:7}
\end{equation}
Here, $a^i_t$ represents the percentage of the event sequence that $M^i_t$ occupies at time $t$. The label $y^i$ is the one-hot label of $x^i$. The value of $a^i_t$ is calculated using Equation \ref{eq:8}. 
\begin{equation}
	a^i_t=\frac{\sum M^i_t}{C\times H\times W}
	\label{eq:8}
\end{equation}
\texttt{EventZoom}  is capable of generating different soft labels across time steps. Although more accurate labels could be obtained, in practice, for computational convenience, all the labels are average.

%

\subsection{A Comparative Discussion }

As shown in Table \ref{tab:method_comparasion}, we summarize the comparison between different methods in terms of maintaining Spatial Integrity(SI) and Temporal Continuity(TC).

The \textit{CutMix}   data augmentation technique, which employs a cut-and-paste strategy for image regions, is effective in conventional image processing scenarios but exhibits limitations when applied to event-based data. \textit{CutMix} involves cutting a region from the image, resulting in the information loss, particularly of the temporal label information related to the cropped area, which may not align with the new context, causing label mismatches\cite{uddin2020saliencymix,kim2020puzzle}.  

The \textit{Mixup}  method functions by directly mixing data at the pixel level between two images, effectively simulating a high dimensional state that lies between them. However, for event data, which captures variations in lighting, this straightforward data mixing method can result in non-authentic lighting conditions, generating data that does not accurately reflect the lighting dynamics of the physical world.  

Despite being designed for event data, \textit{EventMix/ EventRPG/ EventAugmention}, shares spatial issues similar to those of \textit{CutMix}. Specifically, this type of methods not effectively address the consistency of spatial positioning during the cutting process, which can lead to label mismatches.   

\begin{table}[t]
	\centering

	\scalebox{0.70}{
		\setlength{\tabcolsep}{0.2mm} {  
			\begin{tabular}{lccc} 
				\toprule
				\textbf{Model} & \textbf{\makecell[c]{Event\\Augmentation}} &  \textbf{ \makecell[c]{Spatial Integrity\\(SI))} }   	& \textbf{\makecell[c]{ Temporal Continuity\\(TC) } } \\ 
				\midrule
				CutMix	       & \textcolor{black}{\ding{55}}  & \textcolor{black}{\ding{55}} & \textcolor{black}{\ding{55}}         \\
				MixUp	   	   & \textcolor{black}{\ding{55}}  & \textcolor{black}{\ding{51}}  &\textcolor{black}{\ding{55}}         \\
				NDA	   	   & \textcolor{black}{\ding{51}}  & \textcolor{black}{\ding{55}}  &\textcolor{black}{\ding{55}}         \\
				ShapeAug	   	   & \textcolor{black}{\ding{51}}  & \textcolor{black}{\ding{51}}  &\textcolor{black}{\ding{55}}         \\
				EventAugment	   	   & \textcolor{black}{\ding{51}}  & \textcolor{black}{\ding{55}}  &\textcolor{black}{\ding{55}}         \\
				EventDrop	       & \textcolor{black}{\ding{51}}  & \textcolor{black}{\ding{55}} & \textcolor{black}{\ding{55}}         \\
				EventRPG	       & \textcolor{black}{\ding{51}}  & \textcolor{black}{\ding{55}} & \textcolor{black}{\ding{55}}         \\
				EventMix	   & \textcolor{black}{\ding{51}} & \textcolor{black}{\ding{55}} & \textcolor{black}{\ding{55}}         \\
				EventZoom	   & \textcolor{black}{\ding{51}} & \textcolor{black}{\ding{51}} & \textcolor{black}{\ding{51}}         \\
				\bottomrule
	\end{tabular}}}
	\caption{The comparison of whether different methods maintain \textbf{SI} and \textbf{TC} after event data augmentation processing.}
	\label{tab:method_comparasion}
\end{table}












\begin{table*}[h]
	\centering
	
	\scalebox{0.8}{
		\setlength{\tabcolsep}{1.0mm} { 
			\begin{tabular}{lccccc}
				\toprule
				\rowcolor{black!10!white}\textbf{Dataset}    & \textbf{Data Augmentation} & \textbf{Training Method} & \textbf{Neural Network}   & \textbf{Resolution}  & \textbf{Accuracy}  \\ 
				\hline
				& EventDrop\cite{gu2021eventdrop}                  & STBP                   & Pre-Act Resnet18       &   (48,48)                	& 77.73            \\
				& ShapeAug\cite{bendig2024shapeaug}                & STBP                   & Pre-Act Resnet18       &   (128,128)               	& 75.70              \\
				
				& NDA\cite{li2022neuromorphic}                     & STBP-tdBN              & Spike-VGG11             &   (48,48)               & 79.60             \\
				DVS-CIFAR10 
				& NDA\cite{li2022neuromorphic}                     & STBP-tdBN              & Spike-VGG11             &   (128,128)             & 81.70             \\
				& EventMix\cite{shen2023eventmix}              & STBP                   & Pre-Act Resnet18       &   (48,48)               & 81.45             \\
				& EventRPG\cite{sun2024eventrpg}                   & TET                    & Spike-VGG11            &   (48,48)                   & \underline{85.55  }            \\
				& \textbf{EventZoom}         & STBP                   & Pre-Act Resnet18        &   (48,48)                 & 85.40            \\
				& \textbf{EventZoom}         & STBP                   & Spike-VGG11             &   (48,48)              & \textbf{85.90}     \\
				\cmidrule(r){2-6}
				\multirow{8}{*}{N-Caltech101}
				& EventDrop\cite{gu2021eventdrop}                  & STBP                    & Pre-Act Resnet18      &   (48,48)  				    & 74.04              \\
				& ShapeAug\cite{bendig2024shapeaug}                    & STBP                    & Pre-Act Resnet18      &   (128,128)                    & 68.70              \\		    
				& EventAugment\cite{gu2024eventaugment}               & STBP                    & Spike-VGG11            &   (48,48)                   & 75.23             \\
				& NDA\cite{li2022neuromorphic}                         & STBP-tdBN               & Spike-VGG11           &   (48,48)                    & 78.20              \\
				& NDA\cite{li2022neuromorphic}                         & STBP-tdBN               & Spike-VGG11           &   (128,128)                     & 83.70              \\
				& EventMix\cite{shen2023eventmix}                    & STBP                    & Pre-Act Resnet18       &   (48,48)                   & 79.47              \\
				& EventRPG\cite{sun2024eventrpg}                   & TET                     & Spike-VGG11            &    (128,128)             & \underline{85.00}             \\
				& \textbf{EventZoom}         & TET                     & Spike-VGG11            &   (48,48)                   & \textbf{85.75}     \\
				\cmidrule(r){2-6}
				&  C3D\cite{ji20123d}                        & BP                      & ConvNet                 &   (48,48)                     &  47.20                \\ 
				UCF101-DVS			    
				&  EventMix\cite{shen2023eventmix}                   & BP                      & Resnet18-ANN             &   (48,48)                     & \underline{60.63}               \\
				& \textbf{EventZoom}         & STBP                      & Pre-Act Resnet18        &   (48,48)                      & \textbf{62.38}     \\
				\bottomrule
	\end{tabular}}}
	\caption{Comparison of different data augmentation methods across various datasets. Due to the different Settings adopted by different methods, we list the corresponding model architecture, training method, and corresponding resolution.}
	\label{tab:sup_exp}
\end{table*}

\subsection{Checking the Importance of Spatial Integrity and Temporal Continuity}
To demonstrate the significance of spatial integrity and temporal continuity in  data augmentation, we compared the performance disparities under various strategies in our ablation study.\\
1.\textbf{Scaling} can be configured as \textbf{Progressive Scaling(PS)}, \textbf{Random Scaling(RS)}, or \textbf{Static Scaling(SS)}.\\
2.\textbf{Shifting} can be configured as \textbf{Progressive Position(PS)},\textbf{ Random Position(RS)}, or \textbf{Static Position(SP)}.\\
2.\textbf{Cropping} can be configured as either \textbf{Cropping(C)} or \textbf{Scaling}.\\
As illustrated in Figure \ref{fig:different_setting}, different box plots correspond to various settings. Random scaling and shifting disrupt temporal continuity, whereas cropping methods compromise spatial integrity. The comparison highlights the progressive advantages of Eventzoom. More detailed experimental setup and explanation are provided in Appendix B.

\section{Experiment}
To validate the effectiveness of \texttt{EventZoom}, we conducted comprehensive experiments across three distinct event-driven datasets: DVS-CIFAR10, N-Caltech101, UCF101-DVS.   These datasets were rigorously tested in supervised, semi-supervised, and unsupervised learning settings to comprehensively evaluate \texttt{EventZoom}'s performance across varying learning paradigms.
DVS-CIFAR10 is a dynamic vision version of the classic CIFAR10 image dataset, comprising dynamic visual data across ten categories, designed for fundamental image recognition tasks. N-Caltech101, repurposed from the Caltech101 image dataset, includes data for 101 object categories captured by event cameras. UCF101-DVS, derived from the popular UCF101 video dataset, is tailored for action recognition studies and features a diverse array of motion scenarios.  
We used spiking neural networks (SNNs) \cite{wang2020supervised,fang2021deep,zhu2024tcja,wu2021progressive} for validation of each group of experiments. Additionally, to align with current training paradigms and ensure fair comparison with existing methodologies, we adopted the common practice of converting event sequences into frames to facilitate neural network training.


\subsection{Supervised Learning}

We performed extensive experiments to compare \texttt{EventZoom} with other event data augmentation strategies. Table \ref{tab:sup_exp} shows the accuracy of different methods across all datasets. The results clearly indicate that \texttt{EventZoom} significantly outperforms other event augmentation methods. Our approach surpasses the best-performing existing event augmentation methods even  under weaker training setting.

Moreover, we compared \texttt{EventZoom} against traditional image data augmentation methods directly applied to event data. We highlight that, despite the distinct nature of the data types, these methods still demonstrate modest performance improvements when adapted to event data. Experimental results for each dataset are shown in Table  \ref{tab:sup_exp2}. For instance, the mixup method was extended to the temporal dimension, blending corresponding frames between different samples. In the cutmix method, we pasted patches from one frame onto the corresponding position in another sample over time. As a contrast, we also implemented eventmix, which randomly selects Gaussian-distributed samples in the sequence and mixes them with others.

Table \ref{tab:sup_exp2} demonstrates that \texttt{EventZoom} achieved the best performance. Eventmix, compared to CutMix, maintains the spatial integrity, resulting in superior performance. Meanwhile, the MixUp method, which typically preserves structural integrity in neatly structured image data, might disrupt illumination information in event data, leading to poorer performance. \texttt{EventZoom} not only maintains complete spatial information without using patch synthesis by cropping but also preserves full temporal information, increasing diversity without disrupting coherence. DvsGesture and Bullying10K datasets experiments are shown in Appendix.


\begin{table}[ht]
	\centering
	
	\centerline{\scalebox{0.7}{
			\setlength{\tabcolsep}{0.15mm} { 
				\begin{tabular}{ccccc}
					\toprule
					\rowcolor{black!10!white}\textbf{Data Augmentation} & \textbf{DVS-CIFAR10 }	& \textbf{N-Caltech101}  								& \textbf{UCF101-DVS }   \\ 
					\hline
					&\multicolumn{3}{c}{Pre-Act ResNet-18} \\
					\cmidrule(r){2-4}
					No Augmentation                  & \cellcolor{white!00!white}80.80 & \cellcolor{white!00!white}67.93 & \cellcolor{white!00!white}55.54    \\
					MixUp \cite{zhang_mixup_2018}    & \cellcolor{white!10!white}81.40 & \cellcolor{white!10!white}68.62 & \cellcolor{white!00!white}56.36   \\
					CutMix \cite{yun_cutmix_2019}    & \cellcolor{white!00!white}80.70 & \cellcolor{white!00!white}67.93 & \cellcolor{white!10!white}57.58  \\
					\makecell[cc]{EventMix \\ \cite{shen2023eventmix}} & \cellcolor{white!20!white}84.60 & \cellcolor{white!20!white}70.45 & \cellcolor{white!20!white}58.26     \\
					\textbf{EventZoom}        		  & \cellcolor{white!30!white}\textbf{85.40} $_{{\textcolor{black}{(+4.60)}}}$	& \cellcolor{white!30!white}\textbf{78.39}$_{{\textcolor{black}{(+10.46)}}}$	 & \cellcolor{white!30!white}\textbf{62.38}  $_{{\textcolor{black}{(+6.84)}}}$	  \\
					\midrule
					&\multicolumn{3}{c}{Shallow-Spiking-VGG11} \\
					\cmidrule(r){2-4}
					No Augmentation           		  & \cellcolor{white!00!white}81.40 & \cellcolor{white!10!white}71.49 & \cellcolor{white!00!white}50.43          \\
					MixUp\cite{zhang_mixup_2018}	  & \cellcolor{white!10!white}83.60 & \cellcolor{white!00!white}68.96 & \cellcolor{white!00!white}53.91         \\
					CutMix  \cite{yun_cutmix_2019}	  & \cellcolor{white!00!white}81.90 & \cellcolor{white!00!white}70.11 & \cellcolor{white!10!white}55.16           \\
					\makecell[cc]{EventMix \\ \cite{shen2023eventmix}}  & \cellcolor{white!20!white}84.40 & \cellcolor{white!20!white}73.67 & \cellcolor{white!20!white}57.20        \\ 
					\textbf{EventZoom}        		  & \cellcolor{white!30!white}\textbf{84.80}$_{{\textcolor{black}{(+4.40)}}}$	 & \cellcolor{white!30!white}\textbf{80.00}$_{{\textcolor{black}{(+8.51)}}}$	 & \cellcolor{white!30!white}\textbf{63.54}     $_{{\textcolor{black}{(+13.11)}}}$	      \\ 
					\bottomrule
	\end{tabular}}}}
	\caption{Comparison of different conventional data augmentation methods across various datasets. Deeper colors represent higher accuracy levels.}
	\label{tab:sup_exp2}
\end{table}

\begin{table*}[ht]
	\centering
	
	\scalebox{0.8}{
		\setlength{\tabcolsep}{0.5mm} { 
			\begin{tabular}{lcccccc }
				\toprule
				\rowcolor{black!10!white}
				\multirow{1}{*}{ \textbf{Model}}	&\multirow{1}{*}{\textbf{Data Augmentation}}& \multicolumn{3}{c }{DVS-CIFAR10}  \\
				
				& 									 & \textbf{40 labels $\uparrow$ }					  & \textbf{100 labels$\uparrow$}  				   & \textbf{250 labels$\uparrow$ }   \\ 
				\midrule
				& No Augmentation           		 & \cellcolor{white!00!white}57.00 		  & \cellcolor{white!00!white}70.20 		   & \cellcolor{white!00!white}78.90           \\
				& MixUp \cite{zhang_mixup_2018}      & \cellcolor{white!20!white}63.10		  & \cellcolor{white!10!white}73.70 		   & \cellcolor{white!20!white}81.90          \\
				Pre-Act ResNet-18	    
				& CutMix \cite{yun_cutmix_2019}      & \cellcolor{white!10!white}61.90 		  & \cellcolor{white!20!white}74.40 		   & \cellcolor{white!10!white}80.80           \\
				& EventMix \cite{shen2023eventmix}   & \cellcolor{white!00!white}61.00 		  & \cellcolor{white!00!white}72.80 		   & \cellcolor{white!00!white}79.50        \\
				& \textbf{EventZoom}                 & \cellcolor{white!30!white}\textbf{71.60}$_{{\textcolor{black}{(+20.60)}}}$	& \cellcolor{white!30!white}\textbf{76.80}$_{{\textcolor{black}{(+6.60)}}}$	& \cellcolor{white!30!white}\textbf{82.40}       $_{{\textcolor{black}{(+3.50)}}}$	  	 \\
				\cmidrule(r){2-5}
				& No Augmentation           		 & \cellcolor{white!00!white}63.30 		  & \cellcolor{white!00!white}70.00  		   & \cellcolor{white!00!white}76.80  		 \\
				& MixUp  \cite{zhang_mixup_2018}     & \cellcolor{white!10!white}69.60  		  & \cellcolor{white!20!white}75.60  		   & \cellcolor{white!20!white}79.40 				\\
				Shallow-spiking-VGG11  
				& CutMix \cite{yun_cutmix_2019}   	 & \cellcolor{white!20!white}71.20		  & \cellcolor{white!10!white}75.20  		   & \cellcolor{white!10!white}78.70				\\
				& EventMix\cite{shen2023eventmix}    & \cellcolor{white!00!white}68.00		  & \cellcolor{white!00!white}74.10  		   & \cellcolor{white!00!white}78.40 				\\ 
				& \textbf{EventZoom}        		 & \cellcolor{white!30!white}\textbf{71.40}$_{{\textcolor{black}{(+8.10)}}}$	 & \cellcolor{white!30!white}\textbf{76.00}$_{{\textcolor{black}{(+6.00)}}}$	 & \cellcolor{white!30!white}\textbf{80.70} 		$_{{\textcolor{black}{(+3.90)}}}$			\\ 
				\bottomrule
	\end{tabular}}}
	\scalebox{0.75}{
		\setlength{\tabcolsep}{0.5mm} { 
			\begin{tabular}{cc  }
				\toprule
				\rowcolor{black!10!white}
				\multicolumn{2}{c}{Dataset Info} \\
				\midrule
				\textbf{Dataset}						& 	  DVS-CIFAR10     \\
				\midrule
				\textbf{Train portion}					& 	  \textbf{Val portion}	   \\ 
				\midrule
				9000									&	  1000          \\
				\midrule
				\textbf{Used Labels Num	}				&	\textbf{No Labels Num}          \\
				\midrule
				40 $\times$ 10 							&     8600			\\
				100 $\times$ 10 						&     8000			\\
				250 $\times$ 10 						&     6500			\\
				\bottomrule
	\end{tabular}}}
	\caption{Comparison of different data augmentation methods for semi-supervised learning tasks. Experiments were conducted with settings featuring 40, 100, and 250 labels respectively. Deeper colors represent higher accuracy levels.}
	\label{tab:semisup_exp}
\end{table*}

\begin{table}[ht]
	\centering

	\scalebox{0.75}{
		\setlength{\tabcolsep}{0.0mm} { 
			
			\begin{tabular}{ccc}
				\toprule 
				\rowcolor{black!10!white}\multicolumn{1}{c}{\textbf{Model}}&	\multicolumn{2}{c}{Shallow-spiking-VGG11}\\
				
				\midrule
				\textbf{Data Augmentation} & \makecell[ll]{\textbf{P:N-Caltech101/ }\\ \textbf{F:N-Caltech101}}  & \makecell[ll]{\textbf{P:N-Caltech101/ }\\ \textbf{F:DVS-CIFAR10}}  \\ 
				\midrule
				No Augmentation           								& \cellcolor{red!00!white}15.71 		 							 & \cellcolor{red!00!white}25.50        \\
				\makecell[cc]{MixUp \\ \cite{zhang_mixup_2018}}       	& \cellcolor{red!00!white}45.76 $_{{\textcolor{black}{(+30.05)}}}$			 & \cellcolor{red!00!white}37.50 $_{\textcolor{black}{(+12.00)}}$         \\
				\makecell[cc]{EventMix \\ \cite{shen2023eventmix}}   	& \cellcolor{red!00!white}48.28 $_{\textcolor{black}{(+32.57)}}$			 & \cellcolor{red!00!white}39.00  $_{\textcolor{black}{(+17.00)}}$       \\ 
				\textbf{EventZoom}       								& \cellcolor{red!00!white}\textbf{50.00} $_{\textcolor{black}{(+34.29)}}$  & \cellcolor{red!00!white}\textbf{44.10}  $_{\textcolor{black}{(+19.40)}}$       \\ 
				\bottomrule
	\end{tabular}}}
	\caption{Comparison of different data augmentation methods for unsupervised learning tasks.}
	\label{tab:unsup_exp}
\end{table}

\subsection{Semi-supervised Learning}

In semi-supervised learning environments, our experiments focused on conditions of limited labeled data. Assume the dataset is $D$, where the unlabeled part of data is denoted as $D_{unlabel} = \{x_i\}_{i=1}^m$, while the labeled part of data is denoted as $D_{label} = \{x_j\}_{j=1}^n$. 
We employed the architecture from \cite{xie2020unsupervised}, a benchmark commonly used in semi-supervised tasks. In the semi-supervised setting, the efficacy of various data augmentation strategies often depends on how they enhance unlabeled data $\{x_i\}_{i=1}^m$ to adapt the model $M$ to the unlabeled distribution $P\sim{\{x_i\}_{i=1}^m}$. We set up experiments using varying proportions of labeled data to test each augmentation method's ability under conditions of label sparsity. The results of these experiments are summarized in Table \ref{tab:semisup_exp}.

When comparing \texttt{EventZoom} with traditional image augmentations adapted for event data and other event-specific augmentations, \texttt{EventZoom} consistently demonstrated superior performance across all label proportions, particularly in extremely label-sparse settings (e.g., only 40 labeled instances per class). When \texttt{EventZoom} only has 40 labels, it can even have similar results as 100 labels without data augmentation. This improvement suggests that \texttt{EventZoom}'s method of creating complex synthetic event sequences make greater use of the unlabeled data distribution, effectively bridging the gap between labeled and unlabeled data.

\subsection{Unsupervised Learning}

In unsupervised learning environments, our experiments focused on conditions of unlabeled data, defined as $\forall x_i \in D_{unlabel}, \, x_i \in D$.  In particular, we focus on thecontrastive learning algorithms with instance discrimination as a proxy task, which rely heavily on data augmentation. We adopted the architecture in \cite{chen2021exploring}, a benchmark commonly used in unsupervised learning tasks, which models the relationship between the same sample under different augmentations. 
We assessed the model using a linear evaluation, in which, the parameters of the backbone are frozen and append a linear classification layer for learning during this phase. The N-Caltech101 dataset was used as the pre-training dataset. For the linear evaluation, both the DVS-CIFAR10 and N-Caltech101 datasets used as fine-tuning datasets. The results are presented in Table \ref{tab:unsup_exp} .

Without augmentation, the model achieved only low accuracy, making effective learning challenging, highlighting the importance of augmentation for this task. The results under both settings indicate that  \texttt{EventZoom} outperforms traditional augmentation methods and other event-based augmentation techniques. In the Table \ref{tab:unsup_exp}, $P$ represents pre-training and $F$ represents fine-tuning. It is noteworthy that similar unsupervised learning experiments were conducted in NDA. However, in NDA\cite{li2022neuromorphic}, all parameters were used during the linear evaluation phase, whereas we adhered to the standard unsupervised learning paradigm by training only the linear classification layer. This experiment demonstrates that \texttt{EventZoom} effectively utilizes both the temporal and spatial features of event data, significantly enhancing the ability of contrastive learning algorithms to extract and learn robust features from event data.


\begin{table}[t]
	\centering
	
	\centerline{\scalebox{0.7}{
			\setlength{\tabcolsep}{0.5mm} {  
				\begin{tabular}{lccccccc}
					\toprule 
					\rowcolor{black!10!white}\multicolumn{3}{l}{\textbf{Model} }&\multicolumn{2}{l}{ Shallow-Spiking-VGG11 }& \\
					\midrule
					\multicolumn{3}{l}{\textbf{Data Augmentation}}&\multicolumn{2}{l}{ EventZoom} \\
					\midrule
					\textbf{ 0.0 - 0.0 }   &\textbf{  0.2 - 0.6  }	& \textbf{ 0.3 - 0.7 }  &\textbf{  0.3 - 1.0 }&\textbf{  0.4 - 1.3 } &\textbf{  0.5 - 1.5 } \\ 
					\midrule
					\cellcolor{teal!0!white}71.49 & \cellcolor{teal!00!white}71.26 $_{\textcolor{black}{(-0.23)}}$ & \cellcolor{teal!0!white}72.29$_{\textcolor{black}{(+0.80)}}$  & \cellcolor{teal!0!white}74.94 $_{\textcolor{black}{(+3.45)}}$   & \cellcolor{teal!0!white}77.47 $_{\textcolor{black}{(+5.98)}}$& \cellcolor{teal!0!white}80.00  $_{\textcolor{black}{(+8.51)}}$       \\
					\bottomrule
	\end{tabular}}}}
	\caption{Comparison of different $\lambda_{min}$ and $\lambda_{max}$ with model performance. Each value represents the accuracy in the corresponding range ($\lambda_{min}$ - $\lambda_{max}$), where 0.0-0.0 means no data augmentation. }
	\label{tab:ablation_exp1}
\end{table}

\begin{figure}[b]
	\centering
	\centerline{\includegraphics[width=0.9\columnwidth]{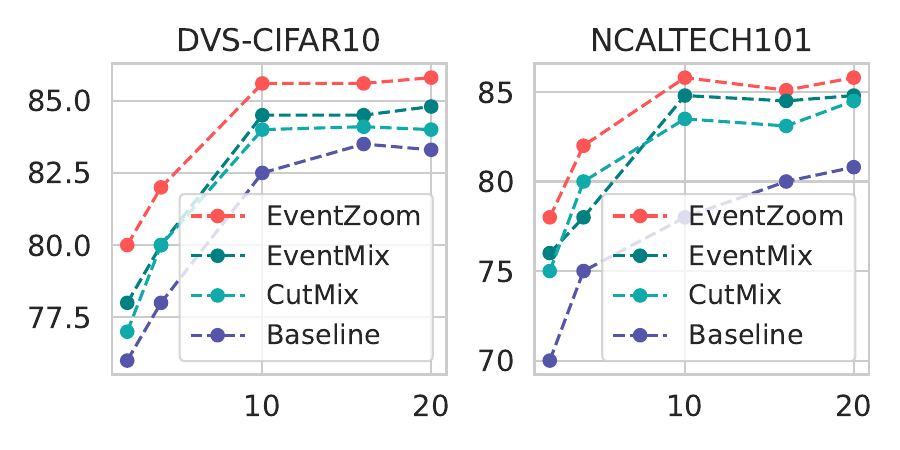} }
	\caption{Comparison of different data augmentation methods at different time steps.}
	\label{fig:different_time}
\end{figure}

\begin{figure*}[h]
	\centering
	\centerline{\includegraphics[width=1.75\columnwidth]{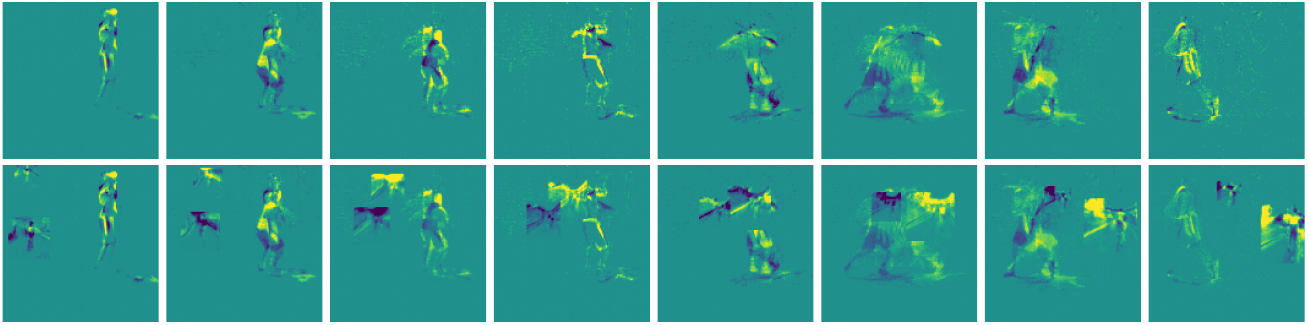} }
	\caption{The figure shows the original (first row) and \texttt{EventZoom} enhanced (second row) version of an event sample. We set $mixnum$ to 2 for illustrative purposes.}
	\label{fig:example}
\end{figure*}

\section{Ablation Study}
To gain a deeper understanding of the impact of various parameters on the effectiveness of the \texttt{EventZoom} data augmentation, we conducted a series of ablation experiments.  To ensure fairness and consistency in testing, we maintained the same settings for all other experimental configurations, except for the parameters used in the ablation studies.

\begin{figure}[t]
	\centering
	\centerline{\includegraphics[width=0.9\columnwidth]{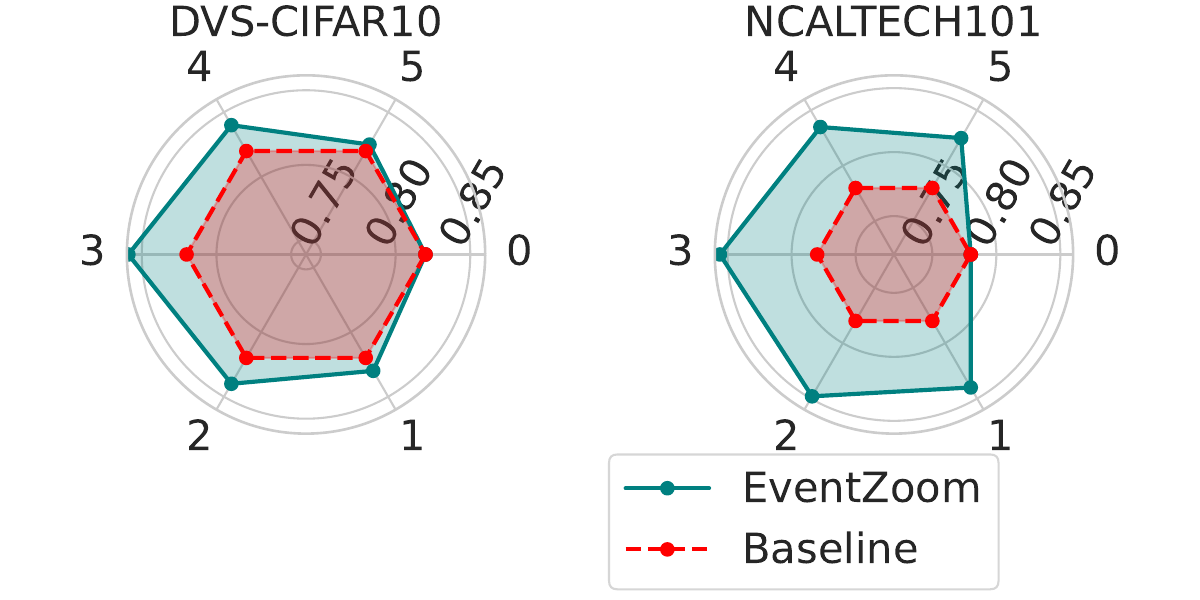} }
	\caption{Radar plot of $mixnum$ with different numbers of embedding samples, where baseline is the benchmark without data augmentation.}
	\label{fig:rader}
\end{figure}

\paragraph{Sensitivity of Different Methods to \textit{Time Step}}
The comparison of different augmentation methods under various time step settings is illustrated in Figure \ref{fig:different_time}. As the \textit{time step} increases, there is a noticeable improvement in model performance. Notably, \texttt{EventZoom} consistently outperforms other methods in accuracy at each time step, showcasing its robust stability across different temporal scales.

\paragraph{Changes in $\lambda_{min}$ and $\lambda_{max}$}
The parameters $\lambda_{min}$ and $\lambda_{max}$ dictate the range of the random scaling factor $\lambda$. Adjusting these parameters enables control over the proportion of embedded event samples. We evaluated \texttt{EventZoom} across various settings of $\lambda_{min}$ and $\lambda_{max}$ to determine the optimal configuration. N-Caltech101 is used in this experiment. As indicated in Table \ref{tab:ablation_exp1}, the model achieved optimal performance within the range of $0.5$ to $1.5$. Notably, a $\lambda$ value greater than $1$ results in data scaling that exceeds the size of the original samples, introducing greater scale diversity.

\paragraph{Illustration of a sample event}
In Figure \ref{fig:example}, we present frame-by-frame visualizations of a sample generated using \texttt{EventZoom}. The augmented effects are visible in this figure. Each frame in the figure is arranged horizontally, with the first row depicting the visualization of the original sample and the second row showing the augmented sample. It is noticeable that the augmented sample exhibits more complex features.

\paragraph{Comparison of the numbers of embedded samples}
We explored how many embedded samples is optimal. The parameter $mixnum$ represents the number of different samples inserted during the augmentation process. As shown in Figure \ref{fig:rader}. Increasing  $mixnum$  may introduce more diversity, but it could also increase the complexity and noise in the training data.  $\lambda_{min}$ and $\lambda_{max}$ are 0.5-1.5. 

\paragraph{Distribution of Labels}
To gain a deeper understanding of \texttt{EventZoom}, we visualized the distribution of labels generated under various $\lambda_{min}-\lambda_{max}$ settings. Notably, \texttt{EventZoom} is capable of producing different labels at different times, which more accurately alleviates the issue of label mismatches.

\begin{figure}[t]
	\centering
	\centerline{\includegraphics[width=1.05\columnwidth]{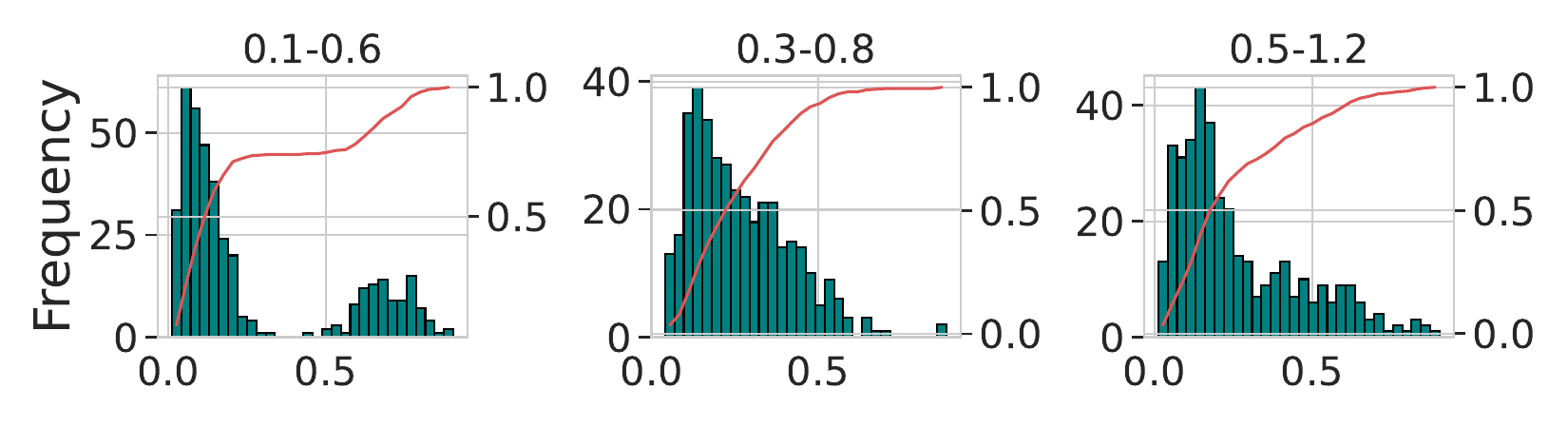} }
	\caption{The distribution of the number of labels under different $\lambda_{min}-\lambda_{max}$. The red line is the cumulative curve.}
	\label{fig:label_distribution}
\end{figure}

A more detailed analysis of the time cost of different augmentation methods is provided in the Appendix.



\section{Conclusion}

This paper proposed a novel event augmentation method \texttt{EventZoom} strategy, which synthesizes new event sequences using a progressive embedding samples technique. \texttt{EventZoom} effectively preserving the \textit{spatial integrity} and \textit{temporal continuity} of the data, which are essential factors affecting the capacity of model, validated and analyzed by specific experiments. We conducted comprehensive experiments of \texttt{EventZoom} across various learning frameworks, including supervised, semi-supervised, and unsupervised learning, highlighting the \texttt{EventZoom} ability by SOTA performance.
Despite these advances, currently \texttt{EventZoom} has only been tested under classification frameworks. Due to the lack of suitable event datasets for downstream tasks, further testing has not yet been conducted. However, our preliminary experiments in downstream tasks still demonstrate the effectiveness of our method.
The development of \texttt{EventZoom}  advances event-based visual processing technology, providing robust support for handling real-world scenes with high dynamics and variability.

\section{Acknowledgments}
This research was financially supported by a funding from Institute of Automation, Chinese Academy of Sciences(Grant No. E411230101)
\bibliography{aaai25}

\newpage
\appendix
\section*{Appendix}
\section{A. Detailed Related Work}
To provide a more comprehensive overview of data augmentation, event-based data augmentation, and the application of data augmentation in semi-supervised and unsupervised learning, we have included a more detailed and complete review of the related work:

\paragraph{Data Augmentation} Data Augmentation is a crucial technique in the field of machine learning, particularly for image processing tasks, as it artificially expands the size and diversity of training datasets. This process helps prevent overfitting, enhances the model's generalization capabilities, and is also robustness to adversarial attack. Traditional data augmentation for image data is categorized into geometric transformations and photometric transformations \cite{maharana_review_2022,shorten2019survey,mumuni2022data}. Geometric transformations include operations like rotation, translation, scaling, and flipping, which alter the spatial structure of the images. Photometric transformations involve adjustments to an image's color, contrast, brightness, and sharpness.
In addition to these basic transformations, several advanced data augmentation techniques have emerged that significantly impact model robustness and training efficiency. For example, the Mixup \cite{mumuni2022data} technique blends features and labels from two or more images to create new samples. Cutout \cite{devries_improved_2017} randomly masks parts of the input image, compelling the network to focus on less prominent features. CutMix \cite{yun_cutmix_2019} combines elements of Mixup and Cutout by replacing part of an image with a segment from another and appropriately blending their labels. Techniques like PuzzleMix\cite{kim2020puzzle} and SaliencyMix\cite{uddin2020saliencymix} further strategically utilize salient regions from different images to enhance the training process. Moreover, automated methods such as AutoAugment\cite{cubuk2019autoaugment} and RandAugment\cite{cubuk2020randaugment} employ reinforcement learning or random searches to discover the most effective augmentation strategies, optimizing them for specific datasets or tasks.

\paragraph{Event Data Augmentation}
The augmentation techniques for event data are specifically designed for their unique attributes, as this type of data, by capturing the dynamic changes in pixel brightness, is closely tied to time. This makes traditional image augmentation methods inapplicable. These event-based techniques need not only capture changes in the imagery but also accurately maintain the temporal information and authenticity of the events. EventDrop\cite{gu2021eventdrop} enhances the regularization ability of the model by randomly dropping events. \cite{li2022neuromorphic} by analyzing the impact of various traditional data augmentation methods on event data and applying them accordingly. EventMix\cite{shen2023eventmix} randomly samples a Gaussian distribution in the event stream and replaces the area with a corresponding area from a random sample. EventRPG\cite{sun2024eventrpg} utilizes CAM and region salience detection techniques to correlate the cut-out area with significant regions. ShapeAug\cite{bendig2024shapeaug} achieves data augmentation by moving simple shapes within the image. EventAugment \cite{gu2024eventaugment} searches for optimal combinations of enhancements through automatic parameter tuning. However, current methods still face issues, such as loss of information during the augmentation process, or complex strategies that make them difficult to apply easily. Our experiments have demonstrated the importance of preserving complete spatial and temporal information for enhancing event data. Moreover, our method is efficient, and introduces minimal additional computational load.

\paragraph{Data Augmentation in Semi(Un)-supervised}
Data augmentation is essential in semi-supervised \cite{berthelot2019mixmatch,sohn2020fixmatch,berthelot2019remixmatch,xie2020unsupervised} and unsupervised learning scenarios\cite{wu2018unsupervised,he2020momentum,chen2020simple,chen2021exploring}, particularly when labeled data is either scarce or completely unavailable. Data augmentation aids models in learning more generalizable features through unlabeled data, as the characteristics hidden in extensive unlabeled datasets more accurately represent the distribution of data compared to limited labeled data. In semi-supervised learning, augmentation techniques effectively expand the limited unlabeled datasets. Techniques such as consistency regularization \cite{xie2020unsupervised} rely heavily on data augmentation. they demand that models maintain consistent predictions across different augmented versions of the same data, thereby significantly boosting the reliability of unlabeled data distribution during training sessions. In the context of unsupervised learning \cite{chen2020simple,chen2021exploring}, where no labels are present, data augmentation is instrumental in learning multiple perspectives of samples and developing robust feature representations. By subjecting input data to various transformations, models can discern the intrinsic structures and variations within the data, which is vital for tasks like contrastive learning. Essentially, data augmentation serves as a form that prompts learning algorithms to concentrate on the data's most informative features, eliminating the need for external labels.

\section{B. Detailed Explanation of Checking the Importance of Spatial Integrity and Temporal Continuity}
Here we provide a more detailed explanation of experiment setting in Figure 1:

To demonstrate the significance of spatial integrity and temporal continuity in  data augmentation, we compared the performance under various strategies in our ablation study. The experiments confirmed that the completeness of spatial and temporal information is essential for effective event data augmentation. For clarity, in our experiments, \textbf{Spatial Integrity(SI)} is defined as the proportion of remaining sparse events in the spatial domain. The sparsity of events allows them to contain more discriminative information compared to RGB images. \textbf{Temporal Continuity(TC)} is defined by whether events change smoothly over time without sudden interruptions or abrupt transitions. This progressive change adds complexity to event augmentation. To assess the impact of these factors on event data, we conducted experiments by excluding these factors individually. NCALTECH101 dataset is used with Spiking-VGG11 model in our experiment.\\
1.\textbf{Progressive Scaling and Position (PS\_PP):} Scaling event samples linearly over time and smoothly shifting positions.\\
2.\textbf{Random Scaling and Random Position (RS\_RP):} Scaling event samples and changing their positions randomly over time, introducing greater complexity and variability in a non-smooth transition manner.\\
3.\textbf{Random Shifting with Fixed Scale (RS\_FP):} Randomly changing the positions of event data samples over time while maintaining a fixed scale.\\
4.\textbf{Random Scaling with Fixed Position (FS\_RP):} Randomly scaling event samples over time without shifting positions.\\
5.\textbf{Fixed Scaling with Fixed Position (FS\_FP):} Fixed scaling event samples over time without shifting their positions.\\
6.\textbf{Cropping with Random Scaling and Fixed Position (C\_RS\_FP):}  Randomly scaling event samples mask over time without shifting their positions. Replacing only the events at the corresponding mask locations.\\
7.\textbf{Cropping with Random Position and Fixed Scale (C\_RS\_FP):}  Randomly scaling event samples mask over time without shifting their positions. Replacing only the events at the corresponding mask locations.\\
8.\textbf{Cropping with Progressive Position and Progressive Scale (C\_PS\_PP):}  Progressive scaling event samples mask over time with Progressively shifting their positions. Replacing only the events at the corresponding mask locations.
As shown in Figure 1, the different box plots correspond to various settings. Maintaining the effect of SI and TC produced the best results, underscoring the crucial importance of  SI and TC .


\begin{figure}[!h]
	\centering
	\centerline{\includegraphics[width=0.85\columnwidth]{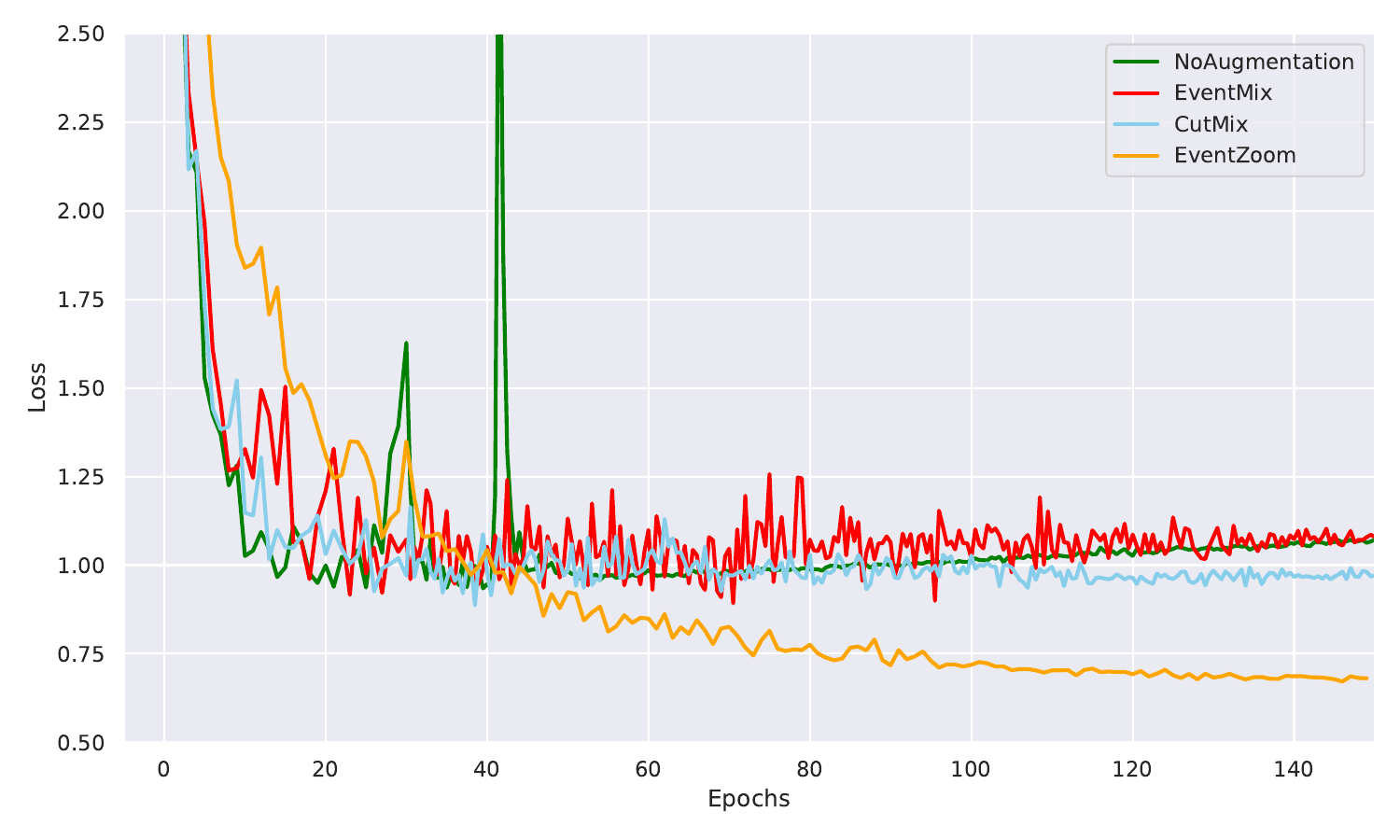} }
	\caption*{Figure A1: Visualization of the Loss curves on the Validation set. The curves of NoAumgentation, EventMix, CutMix, \texttt{EventZoom} are visualized respectively.}
	\label{fig:val_loss_curve}
\end{figure}
\begin{figure}[!h]
	\centering
	\centerline{\includegraphics[width=0.85\columnwidth]{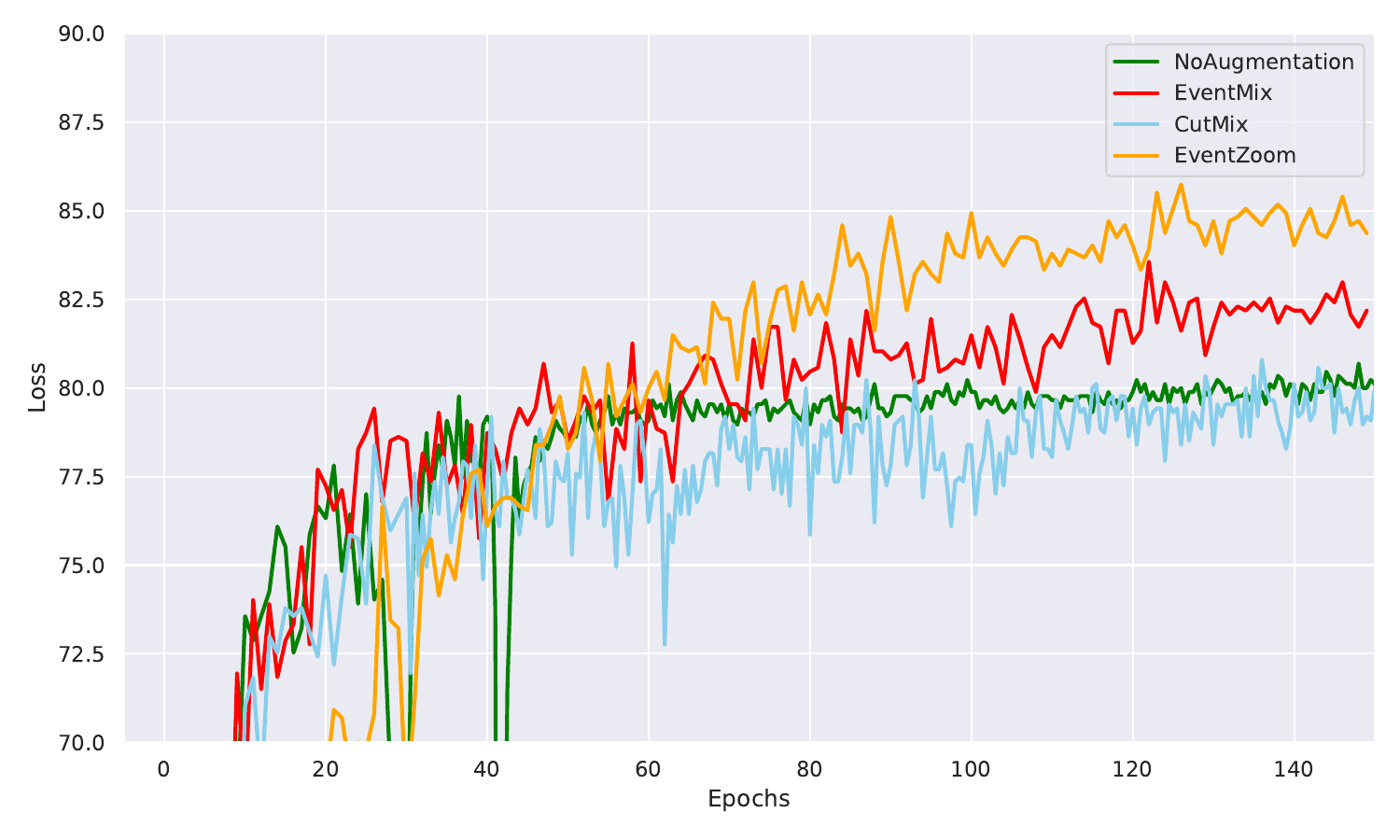} }
	\caption*{Figure A2: Visualization of the Accuracy curves on the Validation set. The curves of NoAumgentation, EventMix, CutMix, \texttt{EventZoom} are visualized respectively.}
	\label{fig:val_acc_curve}
\end{figure}
\begin{figure}[!h]
	\centering
	\centerline{\includegraphics[width=0.85\columnwidth]{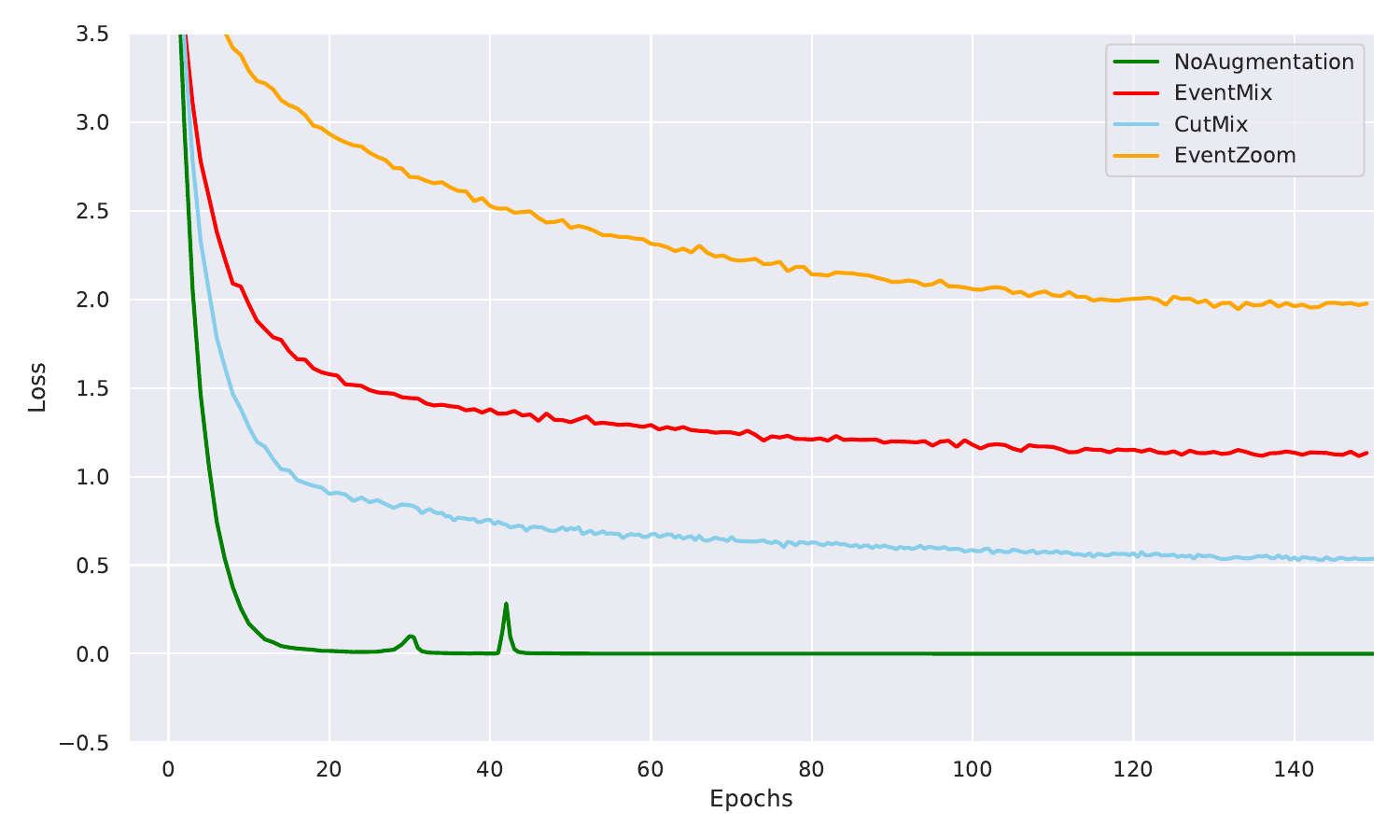} }
	\caption*{Figure A3: Visualization of the Loss curves on the Training set. The curves of NoAumgentation, EventMix, CutMix, \texttt{EventZoom} are visualized respectively.}
	\label{fig:train_loss_curve}
\end{figure}

\begin{algorithm*}[h]
	\caption{\texttt{EventZoom} Event Data Augmentation}
	\label{alg:1}
	\begin{algorithmic}[1]
		\Require Event Dataset $\{S_i\}$, $MixNum$, $\lambda_{min}$, $\lambda_{max}$
		\Ensure Augmented event data
		\State Sample $ori\_sample$ from $\{S_i\}$
		\For {j in range($MixNum$)}
		\State Sample $embed\_sample$ from $\{S_i\}$
		\State Sample $\lambda_s \sim \text{Uniform}(\lambda_{min}, \lambda_{max})$ \Comment{Scaling factor for start time step.}
		\State Sample $\lambda_e \sim \text{Uniform}(\lambda_{min}, \lambda_{max})$ \Comment{Scaling factor for end time step.}
		\State Sample $(x_s, y_s) \sim \text{Uniform}(0, W) \times \text{Uniform}(0, H)$ \Comment{Position for start time step.}
		\State Sample $(x_e, y_e) \sim \text{Uniform}(0, W) \times \text{Uniform}(0, H)$ \Comment{Position for end time step.}
		
		\For {each $t$ from start to end in $embed\_sample$} 
		\State Compute interpolated scale $\lambda_t = (1 - t) \cdot \lambda_s + t \cdot \lambda_e$
		\State Compute interpolated position $(x_t, y_t) = (1 - t) \cdot (x_s, y_s) + t \cdot (x_e, y_e)$
		\State Scale and shift events in time step $t$ of $embed\_sample$ using $\lambda_i$ and $(x_t, y_t)$
		\State Place transformed events in time step $t$ into corresponding position in $ori\_sample$, get $zoomed\_sample$
		\State $y_t =  (C*W*H - Area_t) \cdot y_{ori} + Area_t \cdot y_{embed} $
		\EndFor
		\State $ori\_sample$ = $zoomed\_sample$
		\State $y_{ori}$ = $\{y_t\}$
		\EndFor	
		\State \Return Augmented event data
	\end{algorithmic}
\end{algorithm*}
\begin{figure*}[t]
	\centering
	\includegraphics[width=0.75\columnwidth]{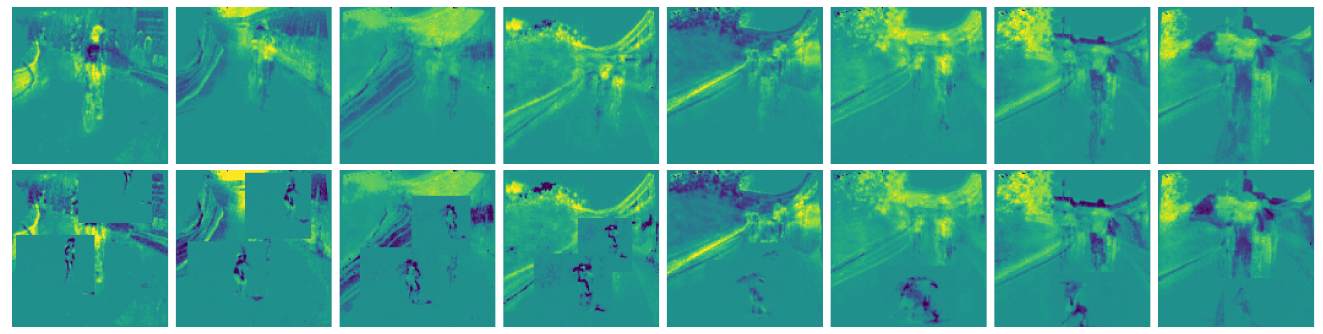} 
	\includegraphics[width=0.75\columnwidth]{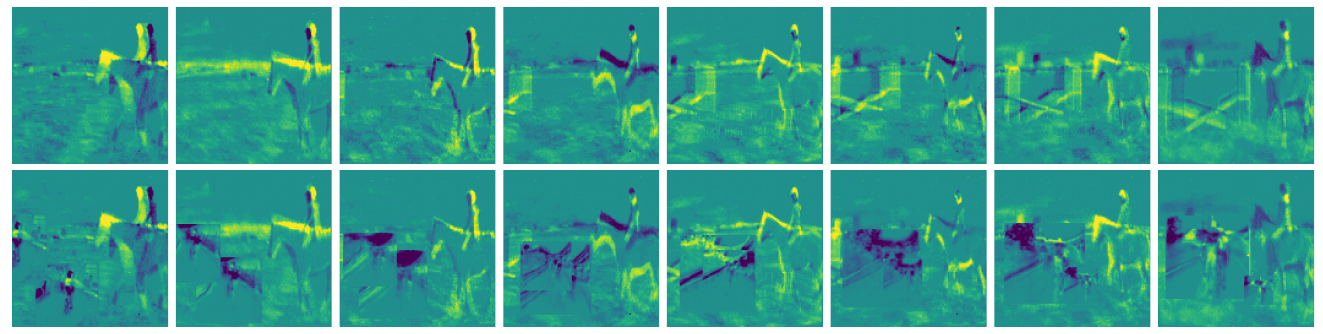}
	\includegraphics[width=0.75\columnwidth]{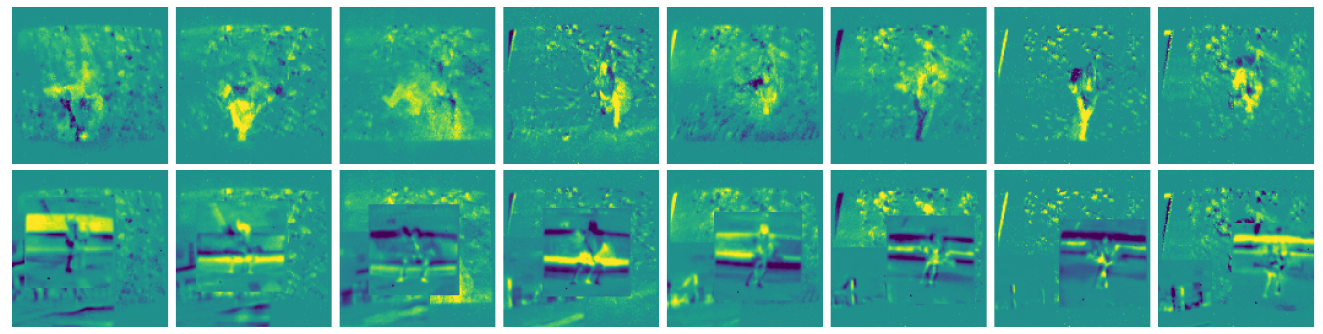} 
	\includegraphics[width=0.75\columnwidth]{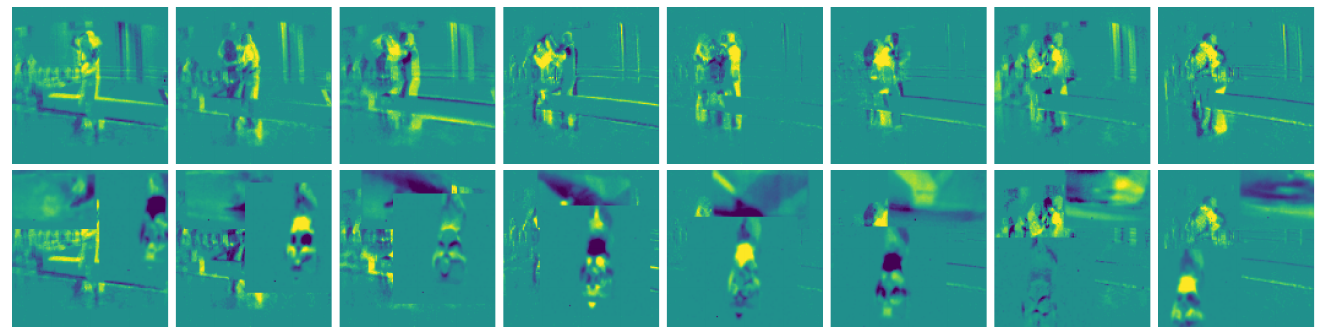}

	\caption*{Figure A4: We visualize more examples  generated by \texttt{EventZoom}, with a time step of 8. The first row contains the original event data and the second row contains the augmented data. }
	\label{fig:example_appendix}
\end{figure*}
\begin{figure*}[!t]
	\centering
	\includegraphics[width=1.8\columnwidth]{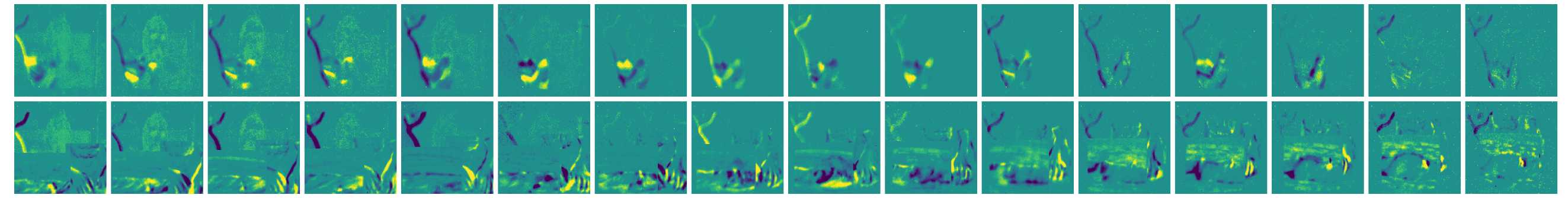} 
	\includegraphics[width=1.8\columnwidth]{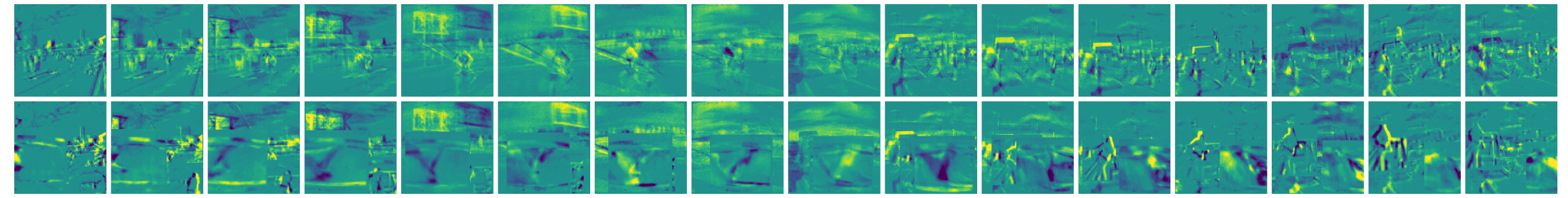}
	\includegraphics[width=1.8\columnwidth]{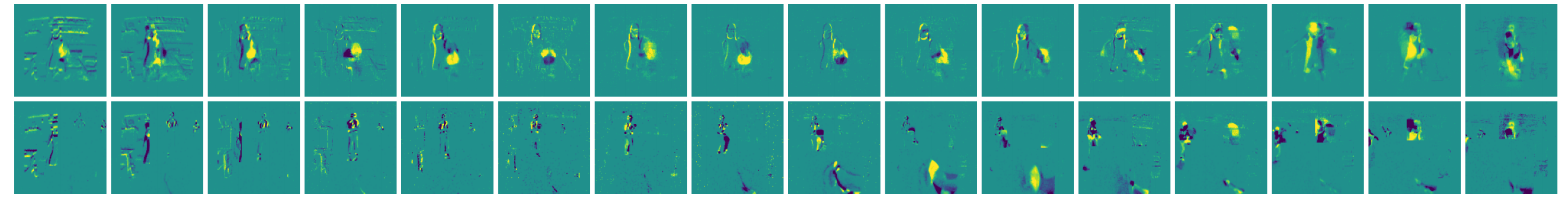} 
	\includegraphics[width=1.8\columnwidth]{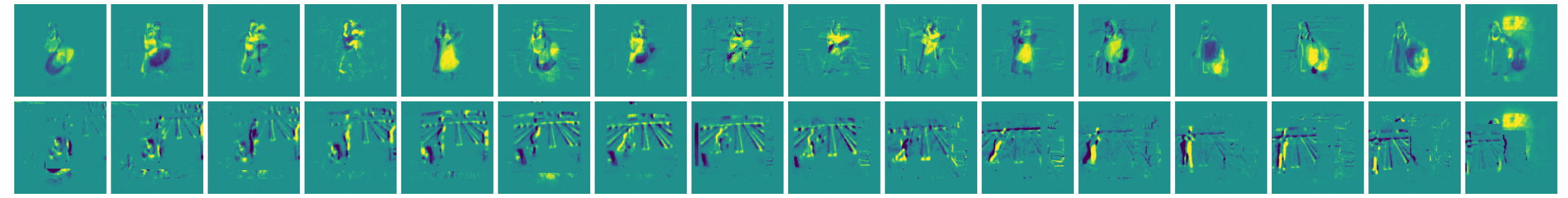} 
	
	\caption*{Figure A5:  We visualize more examples  generated by \texttt{EventZoom}, with a time step of 16. The first row contains the original event data and the second row contains the augmented data. }
	\label{fig:example_appendix2}
\end{figure*}

\section{C. Experimental Environment}  

All our tasks were conducted on a server running the Ubuntu operating system, specifically version 20.04.6. The server is equipped with eight A100 GPUs, each with 40GB of memory. Each experiment was performed on a single A100 GPU. The server is equipped with 503.67GB of memory, and we are using an AMD EPYC 7763 64-Core Processor. We used the PyTorch framework, version 2.3.1, to carry out our experiments. Additionally, we leveraged the BrainCog platform to implement the essential computational components. The same hardware configuration was consistently used across all experiments.

\section{D. Pseudo Code of \texttt{EventZoom}}
To provide a clearer illustration of the EventZoom process, we present the corresponding pseudocode in Algorithm \ref{alg:1}. The pseudocode demonstrates the process of obtaining a single sample from the dataset and applying the EventZoom augmentation to produce a new sample. In other words, we have omitted the handling of the batch dimension in this illustration.

\section{E. Training Curves}
To observe the intermediate states of the model training process under the augmentation of \texttt{EventZoom}, we visualized the accuracy and loss curves during training, including comparisons with NoAugmentation, EventMix, CutMix, and \texttt{EventZoom}. These results are shown in Figures A1, A2, and A3, respectively. We observed that the training loss curve decreases as the training progresses. However, due to the presence of data augmentation, the training loss curve shows a higher loss compared to NoAugmentation. The greater the degree of data augmentation, the higher the loss observed. The validation set curves of Loss and Accuracy indicate that NoAugmentation starts with more instability during the early stages of training, but exhibits smaller fluctuations in the later stages compared to the augmented methods. Notably, \texttt{EventZoom} achieves lower loss and higher accuracy, demonstrating its effectiveness.

\begin{table}[t]
	\centering
	
	\centerline{\scalebox{0.8}{
			\setlength{\tabcolsep}{0.15mm} { 
				\begin{tabular}{ccc}
					\toprule
					\rowcolor{black!10!white}\textbf{Data Augmentation} & \textbf{DvsGesture	 }	& \textbf{Bullying10K}  						   \\ 
					\toprule
					&\multicolumn{2}{c}{Shallow-Spiking-VGG11} \\
					\cmidrule(r){2-3}
					No Augmentation           		  & \cellcolor{white!00!white}94.69 & \cellcolor{white!10!white}73.2         \\
					NDA	  \cite{li2022neuromorphic} & \cellcolor{white!10!white}93.67 & -          \\
					EventRPG \cite{sun2024eventrpg} & \cellcolor{white!10!white}96.53 & -          \\
					CutMix  \cite{yun_cutmix_2019}	  & \cellcolor{white!00!white}94.69 & \cellcolor{white!00!white}71.5        \\
					\makecell[cc]{EventMix \\ \cite{shen2023eventmix}}  & \cellcolor{white!20!white}96.59 & \cellcolor{white!20!white}74.6        \\ 
					\textbf{EventZoom}        		  & \cellcolor{white!30!white}\textbf{96.97}$_{{\textcolor{black}{(+2.28)}}}$	 & \cellcolor{white!30!white}\textbf{80.2}$_{{\textcolor{black}{(+7.00)}}}$	       \\ 
					\bottomrule
	\end{tabular}}}}
	\caption*{Table A1: Comparison of different data augmentation methods across  DvsGesture and Bullying10K datasets. Deeper colors represent higher accuracy levels.}
	
\end{table}

\section{F. Examples}

We provided additional examples to showcase the samples \texttt{EventZoom} generated. As shown in Figure A4, A5.

\section{G. Motivation}
Our approach draws on the idea that in image-based data augmentation, various commonly used image-mixing techniques are inspired by CutMix. These methods highlight the benefit of randomly erasing pixel blocks and replacing them with patches from other images, encouraging the model to focus on diverse features rather than specific, narrow patterns, which in turn improves generalization. For event data, which is inherently temporal, embedding an complete event stream along the time dimension is essential for preserving temporal continuity. This approach supports stronger generalization in models trained on sequential event data.

\section{H. More Results in Supervised Learning}
In our experiments, we included the UCF101DVS dataset, which differs from others in that it contains continuous motion rather than planar shifts. Also, testing on real-world datasets can better demonstrate the effectiveness of our approach. We conducted additional experiments on the DvsGesture\cite{amir2017low} and Bullying10K\cite{dong2024bullying10k} datasets, both of which consist of real-world scenes captured by DVS cameras. As presented in the Table A1, our method achieves more effective results in comparison to other approaches. * denotes the result reported in related papers.

\begin{table}[t]
	\scalebox{0.7}{
		\setlength{\tabcolsep}{0.8mm} { 
			\begin{tabular}{cccccc}
				\toprule
				\rowcolor{black!10!white}
				& \textbf{No Augmentation	} & \textbf{MixUp	 }	& \textbf{CutMix}  		& \textbf{EventMix}  & \textbf{EventZoom}   \\ 
				\midrule
				Cost Time&5.75ms&	8.23ms&	8.44ms&	9.53ms&	8.59ms\\			
				\bottomrule
	\end{tabular}}}
	\caption*{Table A2: Comparison of different data augmentation methods in training cost time.}
	
\end{table}

\section{I. Training Cost}
We conducted training cost analysis to enhances the validation of our method's effectiveness. Since data augmentation primarily affects the data loading phase ( in the transforms stage ) rather than the core training cost of the model itself, we conducted a practical comparison focused on the data loading time for different methods. As shown in the Table A2, our method has minimal impact on data loading.

\end{document}